\documentclass{article}

\usepackage{pdfpages}
\usepackage{natbib}

\usepackage[preprint]{neurips_2024}

\usepackage[utf8]{inputenc} 
\usepackage[T1]{fontenc}    
\usepackage{hyperref}       
\usepackage{url}            
\usepackage{booktabs}       
\usepackage{amsfonts}       
\usepackage{nicefrac}       
\usepackage{microtype}      
\usepackage{xcolor}         
\usepackage{graphicx}

\title{Enhancing Supermarket Robot Interaction: A Multi-Level LLM Conversational Interface for Handling Diverse Customer Intents}

\author{%
  Chandran Nandkumar \\
  Department of Cognitive Robotics\\
  Delft University of Technology\\
  Delft 2628 CD\\
  \texttt{chandran0303.cn@gmail.com} \\
  \And
  Luka Peternel \\
  Department of Cognitive Robotics\\
  Delft University of Technology\\
  Delft 2628 CD\\
  \texttt{l.peternel@tudelft.nl} \\
}

\begin{document}

\maketitle

\begin{abstract}
  This paper presents the design and evaluation of a novel multi-level LLM interface for supermarket robots to assist customers. The proposed interface allows customers to convey their needs through both generic and specific queries. While state-of-the-art systems like OpenAI's GPTs are highly adaptable and easy to build and deploy, they still face challenges such as increased response times and limitations in strategic control of the underlying model for tailored use-case and cost optimisation. Driven by the goal of developing faster and more efficient conversational agents, this paper advocates for using multiple smaller, specialised LLMs fine-tuned to handle different user queries based on their specificity and user intent. We compare this approach to a specialised GPT model powered by GPT-4 Turbo, using the Artificial Social Agent Questionnaire (ASAQ) and qualitative participant feedback in a counterbalanced within-subjects experiment. Our findings show that our multi-LLM chatbot architecture outperformed the benchmarked GPT model across all 13 measured criteria, with statistically significant improvements in four key areas: performance, user satisfaction, user-agent partnership, and self-image enhancement. The paper also presents a method for supermarket robot navigation by mapping the final chatbot response to correct shelf numbers, enabling the robot to sequentially navigate towards the respective products, after which lower-level robot perception, control, and planning can be used for automated object retrieval. We hope this work encourages more efforts into using multiple, specialised smaller models instead of relying on a single powerful, but more expensive and slower, model.
\end{abstract}

\section{Introduction}
In recent times, the presence of robots in our daily lives has increased drastically and they are now capable of working side-by-side with humans to achieve a given objective. The paper in \cite{bloss} explains how collaborative robots improve task efficiency, reduce training times for operators and promise greater safety than their autonomous robot counterparts. Since collaborative robots are a vast and growing field in robotics \cite{Goldberg}, multiple works address the need for different approaches to provide efficient, immersive and aware control. The study in \cite{VILLANI2018248} makes a strong argument for the need to implement intuitive user interfaces, which help reduce operation time and operator errors whilst maintaining situational awareness and user engagement. 

There are multiple options available to interact with collaborative robots. To furnish some examples, \cite{manjum}, \cite{Kragic2018InteractiveCR}, and \cite{takarics2008welding} show different implementations of robot collaboration using vision for a variety of applications like pick-and-place to welding; \cite{app12094379} and \cite{10.3389/frobt.2021.707149}  presents the implementation of Augmented Reality for human-robot collaborative surface treatment and task-level authoring respectively whilst \cite{liu} and \cite{solanes} present the use case of Virtual Reality for the control of robotic manipulators and mobile robots. There are various other means of controlling a robot like eye tracking, pose determination, haptics, facial expressions and more. Furthermore, it is also possible to use multiple such interfaces simultaneously to get more precise and accurate results as seen in \cite{shaif2023visionvoice} and \cite{bai2023whisperwand}. 

With the advancements made in Large Language Models that are capable of processing natural language statements and requests in different languages and complexities in a robust manner, we can now implement interfaces that are capable of accepting and understanding a users exact intentions and needs to provide highly specific and effective results. Furthermore, by connecting with an Automatic Speech Recognition and a Text-to-Speech system we can enable voice-based control that offers other benefits such as being hands-free and highly intuitive. However, the variability in the types of requests in terms of complexity and degree of language processing required implies that a supermarket chatbot must be robust enough to handle both straightforward queries such as asking a specific item's availability, position and price to significantly more open-ended and broader queries regarding high-level intents such as recommendations for a specific dinner or items required for a party. Chatbots built by LLMs are also prone to significant hallucinations and mistakes which influence the degree of trust users can place in these systems \cite{Ma2023Beyondchatbotsllm}. Furthermore, the latency of such systems is often extremely high, affecting their degree of usability. This presents an opportunity to invent a new approach capable of resolving as many problems as possible from above.

Currently, LLMs can be used as interfaces for specific purposes using OpenAI's GPTs powered by the state-of-the-art GPT4 Turbo model \cite{OpenAI2023IntroducingGPTs}. Although GPTs are easy to create and deploy, they suffer from a number of demerits such as high latency, non-flexibility in training models for specific use cases and performance issues of the underlying model.  The main contribution of this work lies in trying to improve upon the limitations of this current state-of-the-art. To achieve this, we propose a novel multi-LLM hierarchical conversational agent capable of responding to all kinds of user queries in a friendly manner. This system is evaluated against a custom-made GPT created with the same data and information provided to our approach using the Artificial Social Agent Questionnaire (ASAQ) across 13 parameters \cite{asaq_questionnaire} in a counterbalanced within-subjects experiment.

Thus this paper aims to answer the following research question: How does our novel multi-LLM conversational agent fare on the Artificial Social Agent Questionnaire (ASAQ) and qualitative evaluation against a custom-built state-of-the-art GPT in a human-factors experiment?

The paper is structured as follows. Section 2 covers the design of the multi-LLM conversational agent along with specific details about the various components of our proposed chatbot. Section 3 presents the experimental methodology. Section 4 presents the results of the evaluation of our proposed system against the state-of-the-art GPT using the ASAQ and qualitative questions. Section 5 discusses the main findings and implications of our study along with the link to robotics. Lastly, Section 6 serves as the conclusion. 
\begin{figure*}[h]
    \centering
    \includegraphics[width=\textwidth]{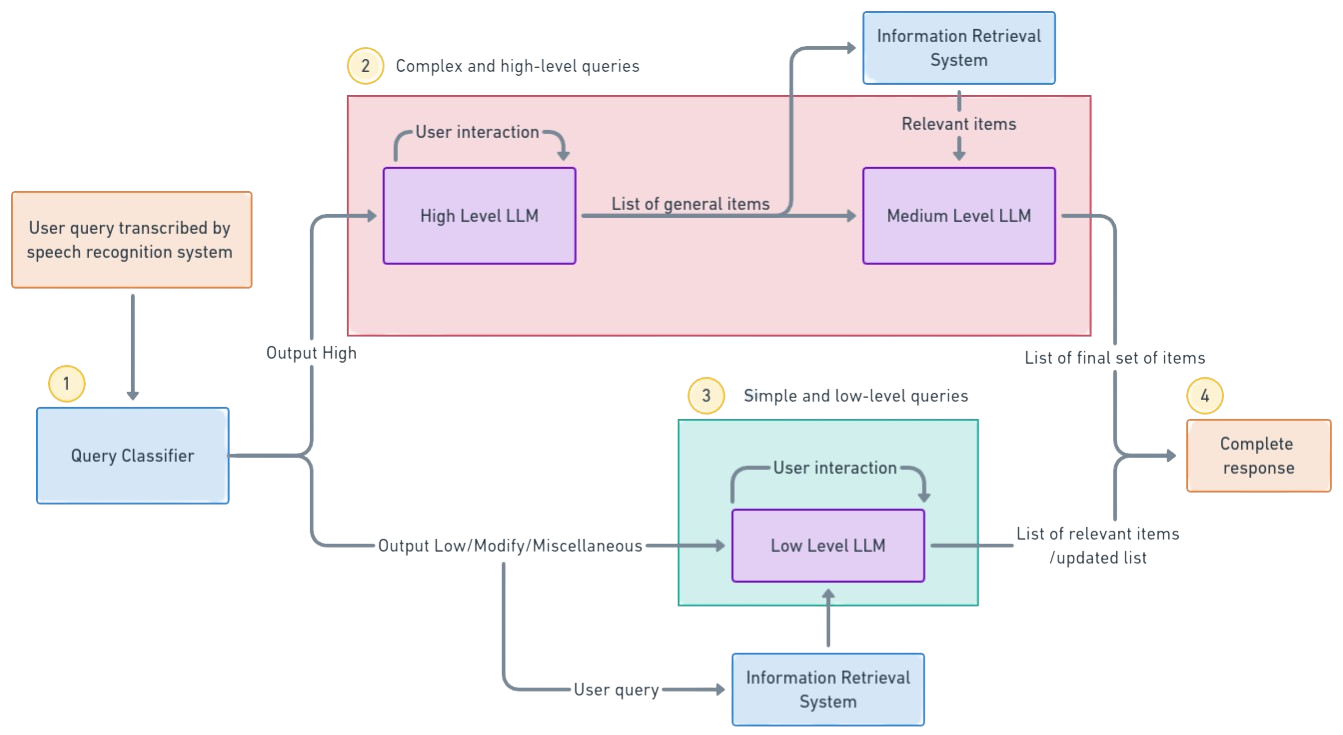}
    \caption{Proposed architecture for handling different queries. Once the query has been transcribed by the speech recognition system, it is classified by the distilBERT system (1). If the query is classified as a high-level query the high-level LLM asks further questions and prepares a rough list of items. These items are sent to the information retrieval system and the relevant items are sent to the medium-level LLM that prepares the correct list of items (2). Otherwise, the query is directly converted to an embedding and searched by the IR system to provide the necessary list of items to the user (3). The relevant response (4) is then shown to the user for further modifications or approval.}
    \label{fig:multillmarch}
\end{figure*}

\section{Design of the Multi-LLM conversational agent} 
\begin{figure*}[h]
    \centering
    \includegraphics[width=\textwidth]{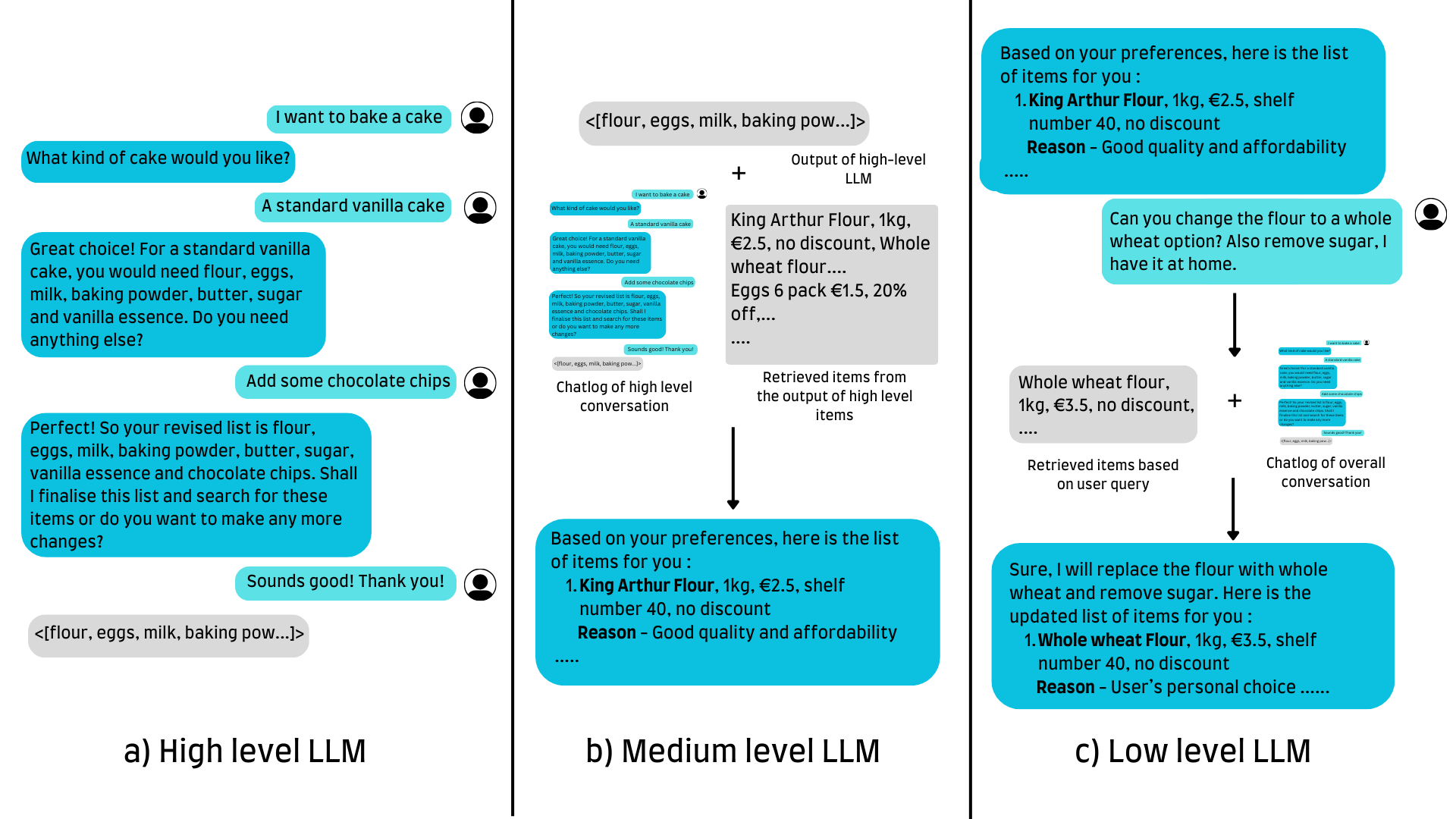}
    \caption{A visual depiction of the responses of the 3 different LLMs. A high-level query takes a request and based on the user's input and user profile, creates a basic list of items. The medium-level LLM takes these items, the chatlog and retrieved items to craft a tailored response for the user. Lastly, all specific queries, modifications and other requests are passed to the low-level query capable of retrieving items and making changes to the original list.}
    \label{fig:hghvsmidvslow}
\end{figure*}

We will begin by covering the main requirements, design strategies and specific details of how we build our supermarket agent. The main requirements of a supermarket chatbot are that it must be able to retrieve relevant information from the database and answer user queries in a friendly and natural manner whilst ensuring it can handle a variety of queries from simple requests asking details about a specific product to complex high-level queries. This requires the conversational agent to not only be capable of basic functions such as natural language understanding, dialogue management and natural language generation but also advanced reasoning and information retrieval.

To address the problems of latency, information retrieval, reasoning window and price, we propose a novel multi-LLM conversational agent where many smaller LLMs, specialised for certain tasks and query types work together to give better results. The architecture we employ is shown in Figure \ref{fig:multillmarch}.

\begin{table*}[ht]
\centering
\begin{tabular}{lcccc}
\toprule
Metric & Accuracy & Precision & Recall & F1 Score \\
\midrule
Value & 0.8679 & 0.8839 & 0.8679 & 0.8651 \\
\bottomrule
\end{tabular}
\caption{Performance Metrics for Set for query classification by DistilBERT}
\label{tab:bert_performance_metrics}
\end{table*}

\subsection{Query Classifier} 

The first step in our conversational agent is to take the input text obtained from the speech recognition system and classify it based on whether the query is high-level, low-level, modification or miscellaneous. High-level queries are those that need to be broken down and analysed with the help of the user to ascertain their preferences, the particular occasion and other restrictions which can enable us to make more informed decisions. A low-level query is a specific request of a particular product or class of products such as finding the location, price or alternatives to an option. A modification query is one where the customer wishes to make amendments to a previously displayed list. Lastly, a miscellaneous statement comprises of everything else such as conversational statements like 'Yes, please' and 'Thank you'.

We need a powerful natural language classifier that can be fine-tuned for the given task. We used distilBERT \cite{sanh2020distilbert} as it can be fine-tuned for our specific classification task and can be run locally on systems without dedicated GPUs. The model is freely available and is easy to train and deploy. The query classifier is trained on over 150 examples primarily augmented by GPT4 by providing a few representational examples to the model. For our fine-tuning purposes, we were able to use anonymous logs of chatbot interactions collected in previous experiments along with GPT4 augmented data. In total, we had 106 English statements, manually labelled from the previous chatlogs and 250 English queries were augmented by GPT4. The data augmentation was done on the ChatGPT interface to allow for better control of the diversity and nature of resultant statements. After shuffling the data, we split the final 356 queries into 250 training, 53 validation, and 53 test sets. Before training the queries were converted to lower case and punctuation marks were removed since we are using a cased distilBERT model. 

The hyperparameters used are as follows - 
\begin{enumerate}
    \item Learning rate : 5e-5,
    \item Number of epochs : 8,
    \item Optimiser : AdamW,
    \item Warmup steps : 10\% of total steps
\end{enumerate}

The final validation loss was 0.58324 and the final validation accuracy was 0.8302 at the 8th epoch and was unchanged from the 7th epoch results. Table summarises the accuracy, recall, precision and F1 scores of the classifier after fine-tuning. The fine-tuned DistilBERT classifier demonstrates a robust performance in classifying queries into four distinct classes: high, low, modify, and miscellaneous, with an accuracy score of 0.8679. The precision score achieved was 0.8839 with a recall score of 0.8679. The F1 Score, which balances precision and recall was found to be 0.8651. It is important to note that the mistakes made in classification are sometimes permissible. For example, the classifier mislabeled 'Sure, add that to my cart.' as 'modify' instead of the ground truth label assigned of 'miscellaneous' which is a completely valid classification for the given query. Likewise, 'I need to replace my usual breakfast cereal with a high fiber option, which one?' was misclassified as a high-level query when the ground truth label assigned was low - which is once again a permissible misclassification since there are multiple options for a high fibre breakfast (high level) but it can also be a low-level query (retrieve the high fibre cereal options).

\subsection{High-level LLM}
If the query is classified as high-level, a high-level LLM is called to interact with the user in order to get more information and break down the query into a list of items the user may need. At this step, user preferences and choices are taken into account along with ascertaining what items the user would need versus that which they already possess or can be substituted. This is best explained with an example. Say the user wishes to bake a cake. There are a number of ingredients needed such as milk, eggs, flour, baking powder, baking soda, vanilla essence, sugar etc. However, the user may possess a lot of these items already at home. Additionally there are other ways to make a cake such as using a cake mix, buying a premade cake or deciding exactly what flavour and nutrition profile you wish to base it on. The high-level LLM is tasked with ascertaining what kind of cake the user wants, if they have any preferences/allergies or other customisations needed along with understanding the exact list of ingredients needed. The high-level LLM is fine-tuned on multi-turn conversations based on prior anonymous chatlogs and few-shot LLM-augmented interactions between a customer and the chatbot.

\subsection{Medium-level LLM}

Once the user is satisfied with this selection of items, the list of user-selected products, the chatlog of the user and the chatbot and the retrieved items are sent to a medium-level LLM that is tasked with creating a tailored list of items from the context with the exact name, brand, price, location and reasoning behind the selection of the items. The medium-level LLM never interacts directly with the user. Based on the response of the medium-level LLM, the user can fine-tune their list of items by making any final changes to the products using the low-level LLM.

The medium-level LLM is also finetuned with various examples of conversations derived from prior anonymous chatlogs and conversations augmented by GPT4. The fine-tuning of this model draws inspiration from Retrieval Augmented Fine Tuning (RAFT) \cite{zhang2024raft}. RAFT provides a simple approach to derive the best of both Retrieval Augmented Generation (RAG) and fine-tuning. An example of its implementation for our application is if 5 different types of flour are retrieved and used as context by the LLM, we specifically use Chain-Of-Thought reasoning to select the whole wheat flour if the user profile indicates that the customer is health conscious. This way, we are not only able to fine-tune our model to present the results in the right format but also can teach it to reason and select the most relevant items from a larger pool of options.

\subsection{Low-level LLM}
Should the user ask for a low-level, modify or miscellaneous query or remark, we call a low-level LLM capable of retrieving the information from the database and giving the output to the user whilst also editing the bill based on the specific request. The process continues until the user is happy with their list and no further edits or changes are necessary. The low-level LLM receives 20 products from the information retrieval system after converting the original query to an embedding and finding the closest neighbours via cosine similarity. 

Similar to the strategy employed in the medium-level LLM, we use RAFT to provide chain-of-thought reasoning during fine-tuning to ensure the correct and most relevant items are picked from the larger pool. Once again the model is trained on both prior anonymous chatlogs and GPT4 augmented conversations to improve its performance for the specific application. 

\section{Experiment Methodology}

Now that we have provided the justification and explanation for our multi-LLM system, we will evaluate our approach against the state-of-the-art GPTs. To do this, we perform a within-subjects experimental study where participants are split into two groups based on the order in which they try both chatbots and are asked to fill our the ASA Questionnaire and provide answers to 3 qualitative questions. 

\subsection{Participant Demographics}
Overall, 16 participants were recruited for the study (9 male and 7 female) between the ages of 23 to 30 (Mean - 24.3125 and SD - 1.8874). In terms of frequency of usage and familiarity with LLM chatbots like ChatGPT, Gemini and Claude, 6 participants responded that they interact with such tools over 5 times a week, 3 responded between 4-5 times a week, 2 responded 3-4 times a week, 4 responded 1-2 times a week and 1 participant responded less than once a week.

\subsection{Experiment Design}
All participants were first shown the informed consent form to ensure that no personally identifiable information will be collected. The only data stored are their responses to the questionnaire, answers to the qualitative questions and chatlogs for further analysis of factors such as hallucinations. We began by collecting demographic details and asking for a brief insight into their shopping intentions such as what they look for and prioritise when they are shopping in the supermarket. Participants were then asked to interact with either the GPT or the custom multi-LLM chatbot we created. The order in which participants tried both chatbots were routinely cycled to ensure half the participants started by interacting with the GPT and the other half with our solution. Participants were not informed of the nature of the agents and were asked to interact with them in a manner they felt best expressed their supermarket intents and goals. 

For the evaluation of our conversational agent, we use the Artificial Social Agent Questionnaire (ASAQ) \cite{asaq_questionnaire}. The questionnaire was developed based on the need to create a validated, standardised measurement instrument dedicated to assessing human interaction with Artificial Social Agents (ASA). The ASAQ is the result of extensive collaboration over multiple years involving over one hundred ASA researchers globally and ensures a robust framework for evaluating interactions between humans and ASAs. The long version of the ASAQ provides an in-depth analysis of human-ASA interactions, catering to comprehensive evaluation needs. Conversely, the short version offers a swift means to analyse and summarise these interactions, facilitating quick insights into the user experience. Additionally, the instrument is complemented by an ASA chart, which serves as a visual tool for reporting results from the questionnaire and provides an overview of the agent's profile. Due to its breadth and comprehensiveness, the ASAQ measures 19 parameters - some of which are not relevant to our study. Thus, 13 relevant parameters were selected which was measured using the short version of the ASAQ.

After interacting with the first chatbot, participants were asked to fill in the 13 relevant questions from the ASAQ followed by the following qualitative questions -
\begin{enumerate}
    \item Tell us in detail, what do you find most helpful and unhelpful from this result.
    \item If at all, how much does this system make you feel more or less confident about your shopping needs and decisions in a supermarket?
    \item Is there anything that you would like to comment about this task?
\end{enumerate}
After this, they were asked to repeat the same procedure but with the other chatbot. The overall experiment took roughly 40 minutes to complete.

Since order is the between-subjects factor and the chatbot is the within-subjects factor, we perform the Mann-Whitney U-test and the Wilcoxon Signed rank test respectively. We use these non-parametric tests since the Shapiro-Wilks test of all the criteria was not normally distributed. This is to be expected given that we were using ordinal data as opposed to continuous values.
\section{Results}

\begin{figure}[ht]
    \centering
    \includegraphics[width=0.8\textwidth]{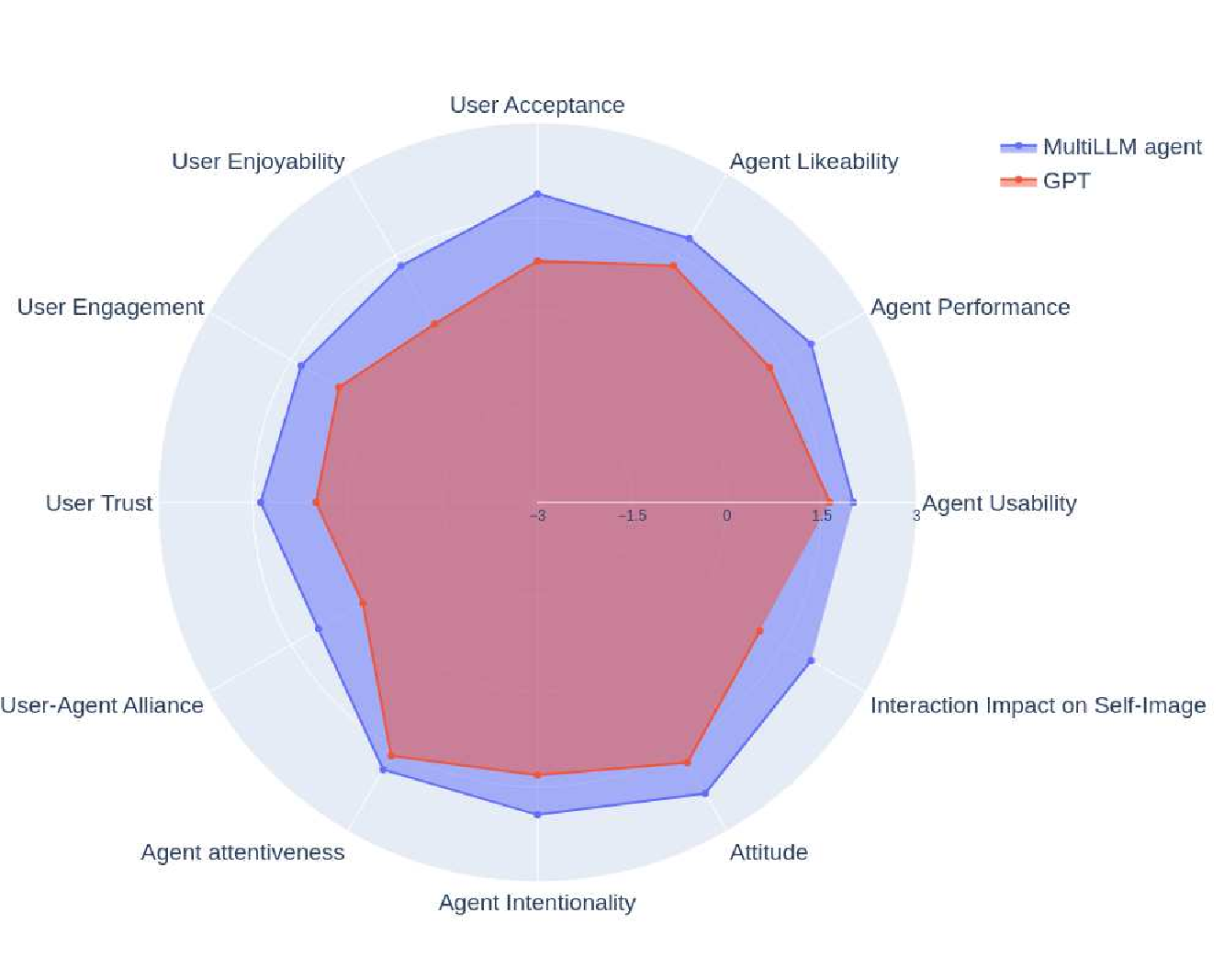}
    \caption{Comparison of the GPT with our custom multi-LLM solution on the provided ASA chart. The scores range from -3 to +3 of the Likert scale on which the ASAQ is built. Our multi-LLM approach performs better than the GPT on all 13 parameters.}
    \label{fig:asaqchart}
    \vspace{2em}
\end{figure}

\begin{table*}[ht]
\centering
\begin{tabular}{@{}clccccccccc@{}}
\toprule
Sl. No & Criterion & \multicolumn{4}{c}{Group Scores (\(\mu\) and \(\sigma\))} & \multicolumn{3}{c}{Statistical Tests} \\
& & GG & GC & CG & CC & Wilcoxon & Mann-Whitney\\ 
\midrule
1 & Agent’s Usability  & 2 & 1.87 & 1.25 & 2.12 & p = 0.19 & p = 0.50 \\
 &  & 0.76 & 0.64 & 1.04 & 0.35 & W = 12.0 & U = 144.5 &  \\
2 & \textbf{Agent's Performance}  & 1.5 & 1.75 & 1 & 2.25 & p = \textbf{0.048} & p = 1.00 \\
 &  & 1.2 & 0.89 & 1.07 & 0.46 & W = 15.0 & U = 128.0 &  \\
3 & Agent’s Likeability  & 1.5 & 1.75 & 1.12 & 1.88 & p = 0.299 & p = 0.814 \\
 &  & 1.41 & 0.71 & 1.46 & 1.13 & W = 36.5 & U = 134.5 &  \\
4 & \textbf{User Acceptance} & 1.12 & 1.75 & 0.5 & 2 & p = \textbf{0.022} & p = 1.00 \\
 &  & 1.13 & 1.04 & 1.85 & 1.31 & W = 13.5 & U = 127.5 &  \\
5 & Agent’s Enjoyability  & 0.25 & 1.25 & 0.25 & 1.38 & p = 0.091 & p = 1.00 \\
 &  & 2.05 & 1.83 & 1.75 & 1.51 & W = 26.0 & U = 127.5 &  \\
6 & User’s Engagement  & 1 & 1.75 & 0.25 & 0.88 & p = 0.095 & p = 0.082 \\
 &  & 1.19 & 0.71 & 0.89 & 1.89 & W = 26.5 & U = 173.0 &  \\
7 & User’s Trust  & 1 & 1.25 & 0 & 1.5 & p = 0.104 & p = 0.63 \\
 &  & 1.31 & 1.04 & 1.51 & 1.77 & W = 22.5 & U = 173.0 &  \\
8 & \textbf{User-Agent Alliance}  & 0.75 & 1.13 & -0.38 & 0.88 & p = \textbf{0.027} & p = 0.065 \\
 &  & 0.71 & 0.83 & 0.92 & 1.73 & W = 0.0 & U = 0.065  &  \\
9 & Agent's Attentiveness  & 1.62 & 1.75 & 1.62 & 2 & p = 0.484 & p = 0.633 \\
 &  & 1.19 & 0.71 & 1.06 & 0.53 & W = 30.5 & U = 115.5 &  \\
10 & Agent's coherence  & 2 & 1.75 & 0.63 & 2.12 & p = 0.108 & p = 0.292 \\
 &  & 0.76 & 1.16 & 1.41 & 0.83 & W = 19.0 & U = 155.0 &  \\
11 & Agent's intentionality  & 2.12 & 2.12 & 1.38 & 2.5 & p = 0.087 & p = 0.732 \\
 &  & 0.64 & 0.99 & 1.06 & 0.76 & W = 6.0 & U = 137.0 &  \\
12 & \textbf{Agent's attitude}  & 1.62 & 2.25 & 0.5 & 1.75 & p = \textbf{0.022} & p = \textbf{0.048} \\
 &  & 0.92 & 0.71 & 1.60 & 0.71 & W = 8.0 & U = 178.0 &  \\
13 & \textbf{Interaction Impact} & 0.88 & 1.5 & 0.25 & 1.5 & p = \textbf{0.017} & p = 0.694 \\
 &  & 0.83 & 1.07 & 1.58 & 0.92 & W = 10.0 & U = 138.5 &  \\
\bottomrule
\end{tabular}
\caption{Summary of Statistical Analysis of the Artificial Social Agent Questionnaire. The 4 groups mentioned are an order model pair and stand for: GG - GPT first GPT scores, GC - GPT first Custom chatbot scores, CG - Custom chatbot first GPT scores and CC - Custom chatbot first custom chatbot scores. All 13 criteria fail the Shapiro Wilks test for normality and thus the Wilcoxon Signed Rank test is done to evaluate the performance between models with p-value and Wilcoxon statistic represented as W and Mann-Whitney U-test is performed to test the effect of order with p-value and Mann Whitney statistic represented as U are presented below.}
\label{tab:asaq_13}
\end{table*}
As seen in Figure \ref{fig:asaqchart} we observe that our solution performs better than the GPT on all 13 tested parameters of the ASAQ. We continue by performing statistical tests on all 13 parameters to find out which parameters are significantly better in our model compared to the state-of-the-art. Table \ref{tab:asaq_13} lists all the 13 parameters. Overall we observe that in terms of agent performance, user acceptance of the agent, user-agent alliance, agent attitude and interaction impact on self-image, the p-value is lesser than 0.05. The Mann-Whitney U-test shows that order is not statistically significant for all criteria except the agent's attitude. Thus we cannot rule out the agent's attitude as being statistically better since order could have influenced the results. 

As mentioned in the experiment design, participants were also asked 3 qualitative questions to try and understand their overall experience better. Participants overall agreed that the state-of-the-art GPT model was simple to use and interact with. Participant \#5 commented on its usefulness as a brainstorming tool to help make decisions of what to purchase and what to try out. Participant \#7 found the responses of the GPT to be more cohesive and in line with their expectations when inquiring about meal preparation strategies for the entire week. Furthermore, participant \#15 found that the responses to complex questions were quite well handled whilst ensuring the conversational tone and language were simple to understand. Whilst none of the participants were overly enthusiastic about the responses and strength of this system, they were content with the answers and recommendations provided by it. However, participants \#2 and \#3 were concerned about hallucinations and mentioned that this affected the degree of trust they could place in the system. P\#2 found some items which did not exist in the database in the responses which were misleading (hallucination) whilst P\#3 was not able to get information about a screwdriver despite the item being present in the database (omission). Participant \#6 had issues substituting organic spinach with regular spinach despite a number of attempts. Participants \#4 and \#8 found the number of options provided by the GPT was limited which made them feel more restricted in terms of choices. Participant \#9 observed that despite mentioning their dietary preferences as being a vegetarian in the user profile, the agent recommended options which did not conform with that. Participant \#14 found that the chatbot was also not able to justify its choices clearly when making recommendations. Multiple participants also commented on the inability of the GPT to provide complete information in its response. For instance, when recommending product names it often forgot to mention the price and location which had to be requested for separately.

Moving on to the custom multi-LLM agent proposed in this paper, participants overall agreed that the proposed chatbot was direct and efficient. Multiple participants commented on the preciseness of the answers which they found made the chatbot very helpful. Although participants were asked to only evaluate the chatbot based on the responses, participants were also impressed with the speed of the chatbot. Participant \#4 commented on how the chatbot reminded them of certain ingredients for their dish that they had forgotten which was very useful. Participant \#5 mentioned that they found the ability to ask questions to narrow down the options to be a helpful feature in the agent. Participant \#7 commented about the reliability and trustworthiness of the agent on account of both the format and reasoning provided by the chatbot. Participant \#11 also mentioned how this chatbot could be useful for people who tend to be more socially anxious and wary of approaching the workers in the supermarket for help and recommendations.

However, participants felt that the chatbot's ability to provide detailed recipes, ideas or plans outside the scope of product recommendation was fairly limited. Participant \#2 stated that they felt the chatbot was more coercive and 'pushy' by trying to force them towards specific products. Participants \#3 and \#5 found that the chatbot made errors when summarising the final list or maintaining track of the conversation. Participant \#8 found that when the LLM was asked to provide the total price of all products, the answer was incorrect. Participant \#13 also commented on how the tool may lead to them purchasing more than they initially sought out.

\section{Discussion}
Overall, we observe that the multi-LLM approach offers multiple benefits over using the most powerful LLM like the state-of-the-art GPT such as reduced costs, reduced latency, increased control over specialised tasks, easier ablation and comparative studies and better task performance. While the GPT solution is indeed the quickest and easiest in terms of deployability, the performance of knowledge retrieval is rather inadequate. By utilising multiple smaller LLMs capable of interacting with one another and maintaining a common conversation log helps in providing context to each separate model as well. The presence of a classifier enables us to directly route queries to the correct model instead of following a common approach for all questions. By fine-tuning with GPT4 augmented data we are also able to leverage the formatting and style of the responses to be extremely well structured and easy to understand.

Furthermore, the modular nature of our solution enables easy substitution of models with alternatives as they become available making the solution extremely flexible to adapt to future developments in the field. One can also fine-tune and use open-source models to ensure reliability and address concerns regarding data privacy and security. Furthermore, by increasing or optimising the number of classes the classifier can select items into, other roles can also be unlocked such as bill management, asking for assistance from supermarket workers or providing feedback. The approach is also not limited to supermarket scenarios and can easily applied to other domains which could benefit from utilising natural language interfaces. By selecting the type of LLMs and queries, the approach can be optimised based on the specific task. 

\begin{figure*}[h]
    \centering
    \includegraphics[width=\textwidth]{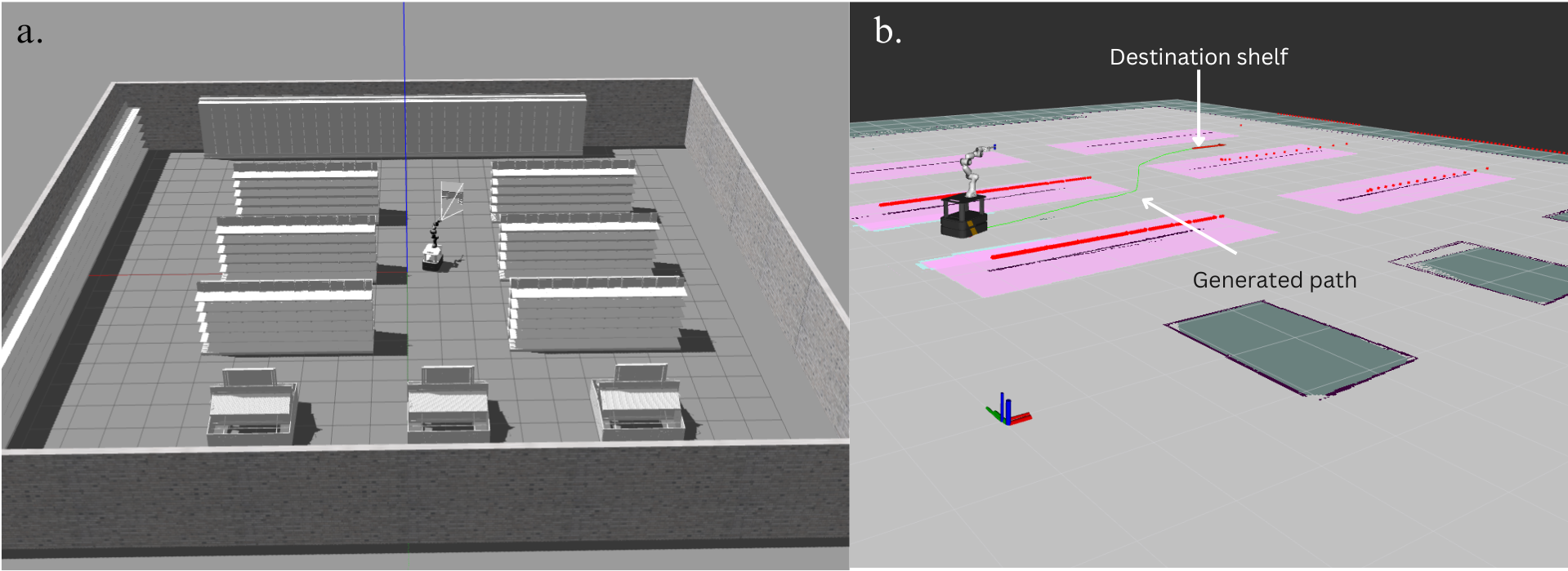}
    \caption{The robot in a large simulated supermarket. Figure a. shows the render on Gazebo while Figure b shows the path (in green line) and the robot navigating to the correct shelf in RViz. The simulation and demonstration have been done on ROS Noetic.}
    \label{fig:robot}
\end{figure*}

While this chatbot can be applied as a standalone application on a mobile phone or in kiosks at the entry of the supermarket, we are also interested in exploring how these chatbots can be effectively integrated into high-level robot planning to guide a supermarket robot to go to the necessary locations after which the required low-level perception, motion planning of a manipulator and control can be applied for automated object retrieval and collection. This feature is useful as it can allow a customer to interact with the chatbot and have a robot autonomously pick up the necessary items and bring them to the user. While the low-level functionality such as perception and manipulation are beyond the scope of our work, we demonstrate with a simple example how our robot can navigate to the necessary shelves after receiving an appropriate request from the customer.

The key assumption made in this work is that the position of all shelves remains the same over time. This is a reasonable assumption to make since most path-planning algorithms require a pre-recorded map to facilitate path planning from a given start point to a destination. If the supermarket is to change its overall configuration, a new map would have to be generated by using SLAM or other similar mapping techniques. To connect the chatbot with the robot, we use an LLM to process the final conversational agent message which has a list of all the products the customer has indicated a willingness to purchase and retrieve a list of shelf numbers for each object. We then define this as a set, removing any duplicates in case multiple items are in the same shelf. The shelves can then be arranged in order to optimise the total distance covered by the robot. We then look up the specific shelf numbers' position from a pre-configured YAML file consisting of the X-Y coordinates of the shelves to retrieve the destination and end pose of the robot. We iterate over all the shelves one after the other until the robot has visited all the necessary items. 

However, the current approach is not without its share of limitations. Incorrectly classified queries can lead to the query being handled by a model that is not specialised for the given task. This could potentially lead to a loss of context and confusing results to users. Since the classifier is built atop a multilingual BERT classifier the responses are highly sensitive to changes in spellings and the manner in which the customer expresses themselves. We believe replacing the mdistilBERT classifier with a small fine-tuned LLM tasked with query classification and rewriting to add any necessary context could be a viable solution to address these limitations and add context to a query to improve retrieval.

\section{Conclusion}
This paper presents the development of a novel multi-LLM agent and evaluates its performance against the state-of-the-art GPT. Our multi-LLM agent surpassed the state-of-the-art in 4 of 13 parameters and demonstrated better performance across all 13 measured ASAQ criteria. The successful integration of LLMs into robot path planning for shelf-directed item retrieval exemplifies the practical application of these interfaces in real-world settings. This study hopes to encourage greater efforts into using multiple specialised LLMs for a required task instead of always relying on a single powerful model to derive benefits in terms of costs, speed and overall task performance.  
\section{Acknowledgements}
The authors would like to thank AIRLab Delft for providing access to the necessary simulation environment and ROS packages to try out the proposed approach. The authors would also like to thank Dr. Chris Pek and Dr. Corrado Pezzato for their insightful feedback on the current limitations and challenges within the field and assistance respectively.

\bibliographystyle{plainnat}
\bibliography{references}
\appendix
\section{Appendix / supplemental material}
\subsection{Participant Instructions}
Participants were welcomed and thanked for participating in the study after which the informed consent form and other details about the experiment and data privacy were communicated. Participants were placed into 2 groups based on the order in which they interacted with the GPT/multi-LLM chatbots. The order was cycled for each subsequent participant. Participants were only informed that they would be interacting with 2 chatbots - one after the other. The overall experiment took roughly 40 minutes per participant.

Before they began, required demographic information was collected such as age, gender, familiarity and frequency of usage of LLMs to ensure that all participants had interacted with such agents to prevent any learning effects from interfering with the study. Next participants were asked to reflect on what their general intentions during a supermarket visit tend to be. The specific instructions provided here were - 

\emph{Please describe, in around 100 words,  any objectives or inquiries you might have while visiting a supermarket. This could range from searching for particular items, seeking advice or recommendations, to any general queries you often find yourself pondering amidst the aisles. Feel free to reflect on your personal needs, preferences, or a specific list of items you aim to purchase.}

\emph{A useful way to approach this is to think about the types of products or goods that usually draw your interest, or those you suddenly remember you need once you're there. Your input can draw upon both past shopping experiences or current needs.}

This question was asked to help participants mentally prepare themselves by thinking of what they would potentially look for in a supermarket so that they could interact with the chatbot more naturally. After this, they were provided with the first chatbot and asked to interact with it until they felt satisfied with the results or enough to make an assessment. They were then asked to fill out the relevant questions in the ASAQ questionnaire. Following this, 3 qualitative questions were provided to which the participants could write their views in whatever depth they deemed necessary. The same process was then repeated for the other chatbot.

\subsection{Remaining Appendices}
All the remaining appendices have been attached as a separate pdf on the next page from A to E.

\includepdf[pages=-, scale=1, pagecommand={}, frame=false]{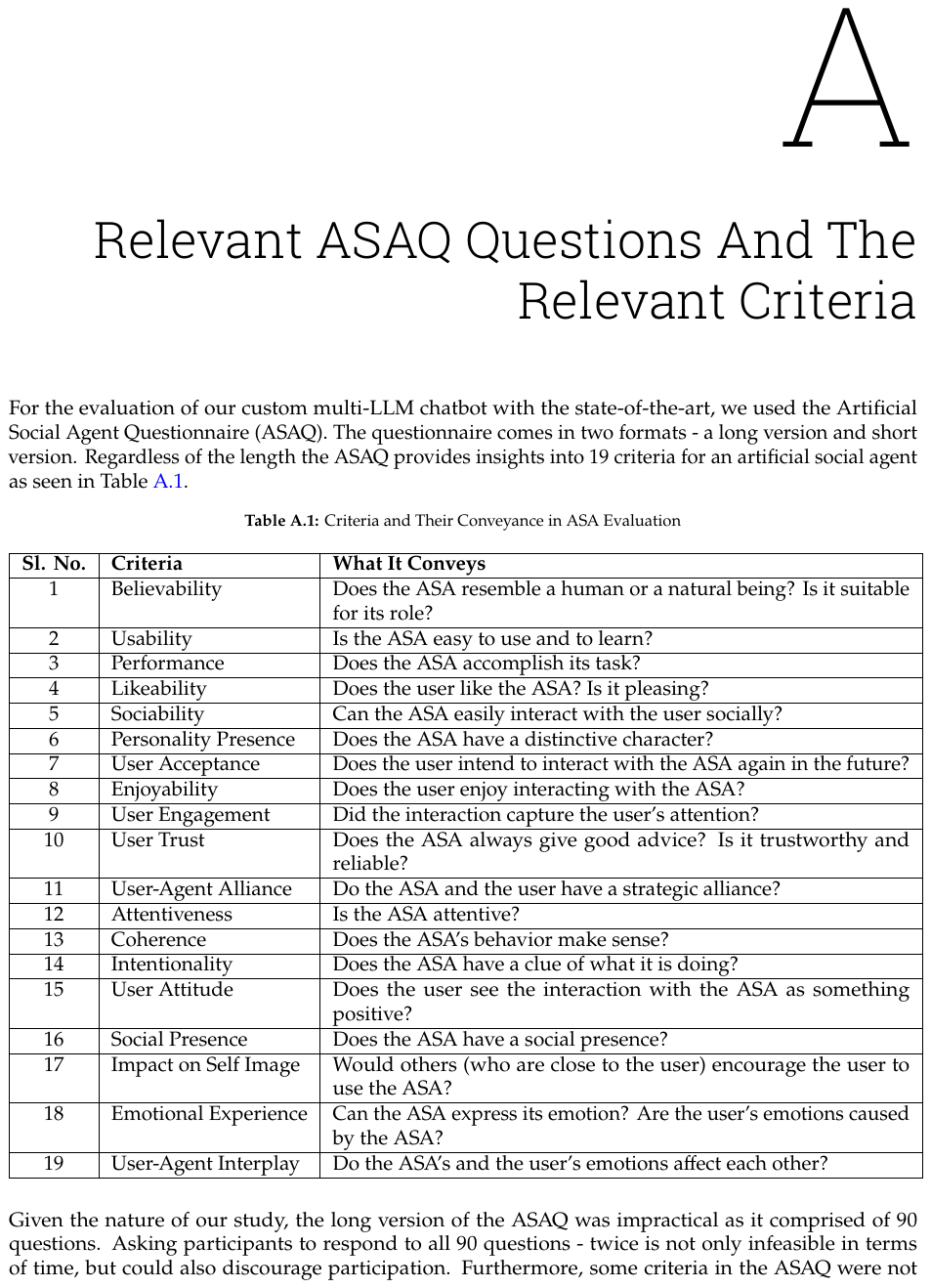}

\newpage
\section*{NeurIPS Paper Checklist}

\begin{enumerate}

\item {\bf Claims}
    \item[] Question: Do the main claims made in the abstract and introduction accurately reflect the paper's contributions and scope?
    \item[] Answer: \answerYes{} 
    \item[] Justification: The paper aims to build and evaluate a novel multiLLM chatbot that is reflected in the abstract and introduction.

\item {\bf Limitations}
    \item[] Question: Does the paper discuss the limitations of the work performed by the authors?
    \item[] Answer: \answerYes{}
    \item[] Justification: The limitations are provided in the last paragraph of the discussions section (Section 5).

\item {\bf Theory Assumptions and Proofs}
    \item[] Question: For each theoretical result, does the paper provide the full set of assumptions and a complete (and correct) proof?
    \item[] Answer: \answerNA{} 
    \item[] Justification: The paper focuses on the design and evaluation of a multi LLM agent and its practical application with no new theoretical proposal. Thus there are no formal proofs in the paper.

    \item {\bf Experimental Result Reproducibility}
    \item[] Question: Does the paper fully disclose all the information needed to reproduce the main experimental results of the paper to the extent that it affects the main claims and/or conclusions of the paper (regardless of whether the code and data are provided or not)?
    \item[] Answer: \answerYes{} 
    \item[] Justification: The full architecture of the approach is provided, the participant demographic information, experiment methodology, evaluation questionnaire and criterion shared and hyperparameters for the classifier are provided. In the Appendix you can also find access to the link for the globally available custom built GPT that can be tested by everyone. 

\item {\bf Open access to data and code}
    \item[] Question: Does the paper provide open access to the data and code, with sufficient instructions to faithfully reproduce the main experimental results, as described in supplemental material?
    \item[] Answer: \answerNo{} 
    \item[] Justification: The code has not been shared as it requires access to some robot design files, robotic worlds and datasets of different products along with their prices for complete execution. This information is sensitive and has been granted as a part of some collaborations with the university and cannot be provided freely. 

\item {\bf Experimental Setting/Details}
    \item[] Question: Does the paper specify all the training and test details (e.g., data splits, hyperparameters, how they were chosen, type of optimizer, etc.) necessary to understand the results?
    \item[] Answer: \answerYes{} 
    \item[] Justification: The hyperparameters for the query classifier are shared in Section 2, the prompts for each model and state-of-the-art are shared in the Appendix.

\item {\bf Experiment Statistical Significance}
    \item[] Question: Does the paper report error bars suitably and correctly defined or other appropriate information about the statistical significance of the experiments?
    \item[] Answer: \answerYes{} 
    \item[] Justification: The standard deviation, mean and statistical test results by both the Mann-Whitney U-test and Wilcoxon signed rank test have been provided.

\item {\bf Experiments Compute Resources}
    \item[] Question: For each experiment, does the paper provide sufficient information on the computer resources (type of compute workers, memory, time of execution) needed to reproduce the experiments?
    \item[] Answer: \answerNo{} 
    \item[] Justification: While specific compute information is not shared, in section 2.1, query classifier we mention how all the experiemnts are done on distilBERT so that it can be run on a system without dedicated graphics card.
    
\item {\bf Code Of Ethics}
    \item[] Question: Does the research conducted in the paper conform, in every respect, with the NeurIPS Code of Ethics \url{https://neurips.cc/public/EthicsGuidelines}?
    \item[] Answer: \answerYes{} 
    \item[] Justification: The informed consent of all participants was obtained as well in line with ensuring the protection of their rights. The informed consent is also provided in the Appendix.

\item {\bf Broader Impacts}
    \item[] Question: Does the paper discuss both potential positive societal impacts and negative societal impacts of the work performed?
    \item[] Answer: \answerYes{} 
    \item[] Justification: In discussion section we explain how our work can be used alongside robots and how the multiLLM approach can be extended to other applications as well.
    
\item {\bf Safeguards}
    \item[] Question: Does the paper describe safeguards that have been put in place for responsible release of data or models that have a high risk for misuse (e.g., pretrained language models, image generators, or scraped datasets)?
    \item[] Answer: \answerNo{} 
    \item[] Justification: Given the highly specific nature of the models and that they are fine-tuned GPT3.5s necessary safeguards were not established. Furthermore, the work does not present any new model but a chat finetuned model finetuned once more for a specific application.

\item {\bf Licenses for existing assets}
    \item[] Question: Are the creators or original owners of assets (e.g., code, data, models), used in the paper, properly credited and are the license and terms of use explicitly mentioned and properly respected?
    \item[] Answer: \answerYes{} 
    \item[] Justification: The GPTs by OpenAI are properly cited and other models used are addressed as well. The acknowledgements are provided in the Appendix.

\item {\bf New Assets}
    \item[] Question: Are new assets introduced in the paper well documented and is the documentation provided alongside the assets?
    \item[] Answer: \answerNA{} 
    \item[] Justification: No new assets are released by the paper.

\item {\bf Crowdsourcing and Research with Human Subjects}
    \item[] Question: For crowdsourcing experiments and research with human subjects, does the paper include the full text of instructions given to participants and screenshots, if applicable, as well as details about compensation (if any)? 
    \item[] Answer: \answerYes{} 
    \item[] Justification: The participant instruction and infored consent form are provided in the Appendix.

\item {\bf Institutional Review Board (IRB) Approvals or Equivalent for Research with Human Subjects}
    \item[] Question: Does the paper describe potential risks incurred by study participants, whether such risks were disclosed to the subjects, and whether Institutional Review Board (IRB) approvals (or an equivalent approval/review based on the requirements of your country or institution) were obtained?
    \item[] Answer: \answerYes{} 
    \item[] Justification: The participant instruction and infored consent form are provided in the Appendix which has information about the risks of the experiment.

\end{enumerate}
\end{document}